# A Vocabulary-Free Multilingual Neural Tokenizer for End-to-End Task Learning

**Md Mofijul Islam**[†,*], **Gustavo Aguilar**[‡], **Pragaash Ponnusamy**[‡]
**Clint Solomon Mathialagan**[‡], **Chengyuan Ma**[‡], **Chenlei Guo**[‡]
University of Virginia [†], Amazon.com [‡]
mi8uu@virginia.edu, {gustalas, ponnup, matclint, mchengyu, guochenl}@amazon.com

## Abstract

Subword tokenization is a commonly used input pre-processing step in most recent NLP models. However, it limits the models' ability to leverage end-to-end task learning. Its frequency-based vocabulary creation compromises tokenization in low-resource languages, leading models to produce suboptimal representations. Additionally, the dependency on a fixed vocabulary limits the subword models' adaptability across languages and domains. In this work, we propose a vocabulary-free neural tokenizer by distilling segmentation information from heuristic-based subword tokenization. We pre-train our character-based tokenizer by processing unique words from multilingual corpus, thereby extensively increasing word diversity across languages. Unlike the predefined and fixed vocabularies in subword methods, our tokenizer allows end-to-end task learning, resulting in optimal task-specific tokenization. The experimental results show that replacing the subword tokenizer with our neural tokenizer consistently improves performance on multilingual (NLI) and code-switching (sentiment analysis) tasks, with larger gains in low-resource languages. Additionally, our neural tokenizer exhibits a robust performance on downstream tasks when adversarial noise is present (typos and misspelling), further increasing the initial improvements over statistical subword tokenizers.

## 1 Introduction

Subword tokenization methods, such as BPE (Sennrich et al., 2016), Word-Piece (Schuster and Nakajima, 2012), and Unigram (Kudo, 2018), rely on a predefined vocabulary to tokenize text. This vocabulary is built based on frequencies of word fragments. As a result, rare words are highly fragmented into many subpieces, whereas the integrity of the most frequent words is substantially preserved (Bostrom and Durrett, 2020). This vocabulary bias is magnified in multilingual settings, where low-resource languages are heavily discriminated in favor of high-resource ones (Tay et al., 2021; Chung et al., 2020; Wang et al., 2021) (see Table 1). Additionally, a subword vocabulary is often defined while processing a (large) pre-training corpus, thereafter remaining fixed. Consequently, when the data samples are drawn from a different distribution (e.g., multilingual text vs. linguistic code-switching, formal writing vs. arbitrary spellings, or simply by adversarial manipulation), the subword tokenizers struggle to adapt and poorly segment the input, in some cases defaulting to character pieces. These issues usually get reflected in downstream tasks when using pre-trained models that rely on subword tokenization (Devlin et al., 2019). The models cannot adapt their predefined static vocabulary, thereby employing suboptimal tokenization for downstream tasks (Clark et al., 2021). We argue that this represents an important bottleneck in the NLP pipeline, where models could become truly end-to-end, but they lag behind due to the only not-learnable component.

---
*Work performed as summer intern at Amazon Alexa AI.

Table 1: Segmentation of *Workshop* in different languages. Subword tokenizers over-segment low-resources languages (Arabic and Thai) and create junk tokens, whereas our neural tokenizer reduces the junk tokens.

| Tokenizers | Word Languages | | |
|---|---|---|---|
| | Arabic | Thai | English |
| BPE | رو / ةش / لم ع | การ/ประชุม/เชิง/ปฏิบัติ/การ | workshop |
| Unigram | شرو / ة / لم ع | การประชุม/เชิง/ปฏิบัติการ | work/shop |
| Word-piece | رو / ةش / لم ع | การ/ประช/ุม/เชิง/ปฏิบัติ/การ | workshop |
| Neural | ةشرو / لم ع | การประชุมเชิงปฏิบัติการ | workshop |

To address the aforementioned issues, we design a vocabulary-free neural tokenizer, which we train in two phases. First, in the pre-training phase, we train our neural tokenizer by distilling the segmentation information from a subword tokenizer. In the multilingual setting, our neural tokenizer learns from the language-specific subword tokenizers so that it is not biased towards high-resource languages. After the pre-training phase, the neural tokenizer segments the character sequence without requiring a predefined vocabulary. In the second phase, we employ an end-to-end learning approach, which allows our neural tokenizer to adapt the tokenization behavior to the downstream task. Such an end-to-end approach is not feasible for models with subword tokenizers due to the predefined vocabulary and their strong ties to the models' embedding layer. Additionally, unlike the subword tokenizers, our neural tokenizer does not require a vocabulary, and its versatile alphabet reduces the bias towards high-resource languages (i.e., there is not unbalanced word coverage favoring specific languages).

We compare the impact of our approach with respect to the subword tokenizers in downstream monolingual, multilingual, and code-switching tasks. For multilingual NLI, the results show that our neural tokenizer generally improves the model performance, with substantially larger gains for low-resource languages (+11 absolute points of accuracy for Thai, +8 for Arabic, and +4 for Swahili). For code-switched Spanish-English sentiment analysis, our neural tokenizer also outperforms the baseline tokenizers, demonstrating better language generalization capabilities. We inspect the robustness of our neural tokenizer in the presence of noisy text through adversarial manipulation (e.g., typos and spelling variations), and we find that the tokenization result is much more resilient to generate *junk tokens* (i.e., excessive fragmentation of subword pieces) than the subword tokenizers. Finally, we provide extensive experimental analyses that consistently suggest to adopt our approach for more robust and versatile representations of text.

## 2 Related Work

### 2.1 Subword Tokenization

Several subword tokenization approaches have been proposed to segment the input text in the NLP pipeline, such as BPE (Sennrich et al., 2016), Word-Piece (Schuster and Nakajima, 2012), Unigram (Kudo, 2018), and SentencePiece (Kudo and Richardson, 2018). These tokenizers use a frequency-based approach to determine the vocabulary from a corpus. Although these subword tokenization approaches improve upon previous rule-based methods, recent studies show that subword tokenization leads the model to produce suboptimal representations (Bostrom and Durrett, 2020; Wang et al., 2021; Chung et al., 2020; Kudo, 2018). For instance, Bostrom and Durrett (2020) evaluate the impact of Byte Pair Encoding (BPE) tokenization on language model pretraining, and the results suggest that BPE leads to suboptimal representations. Due to the data imbalance among the languages, the impact of multilingual tokenization on the representations is profound (Tay et al., 2021; Wang et al., 2021)—i.e., the tokenizers are prone to excessive fragmentation of subwords due to the lack of word coverage leading to meaningless tokens.

To reduce the undermining effects of subword tokenization, several approaches have been proposed. For example, Kudo (2018) introduced a subword regularization approach to probabilistically sample multiple segmentations to improve neural machine translation models. Along this line, Wang et al. (2021) shows that multilingual representations can be improved by utilizing multiple input segmentations. Although these approaches improve the model's representations by using multiple subword segmentation, they ultimately rely on heuristic-based subword tokenization with a fixed vocabulary. Thus, the limitations of the heuristic-based tokenization still persist, such as restricting the model's ability to leverage end-to-end task learning while adapting to an optimal downstream tokenization.

### 2.2 Character-level Models

Although subword tokenization alleviates the out-of-vocabulary problem, it relies on a static vocabulary, which prevents end-to-end learning. A natural alternative to that deficiency is to replace the subword tokenization with a character-level approach and learn the representations directly from the character sequence (Graves, 2013; Sutskever et al., 2011; Radford et al., 2017). These character-based approaches can adapt more easily to noisy text, code-switched languages, and adversarial manipulation to extract the representation (Clark et al., 2021; Tay et al., 2021; Hwang and Sung, 2017; Pinter et al., 2019; Akbik et al., 2018; Xie et al., 2018; Aguilar et al., 2020b). However, the character-based approaches may not capture the token-level representation, which degrades downstream task performance. Moreover, these method have to process longer sequences at the character level, thus increasing quadratically the complexity of the models (Clark et al., 2021; Aguilar et al., 2020b; Costa-jussà and Fonollosa, 2016).

Several approaches have been proposed to downsample the character sequence to sub-token sequence (Tay et al., 2021; Clark et al., 2021). For example, Clark et al. (2021) deterministically combined a fixed number of characters' representations to reduce the model complexity. Along this line, Tay et al. (2021) downsampled the sequence of character vectors by a fixed factor to produce latent subwords representations. Furthermore, Zhang et al. (2019) produced character n-grams, which are hashed and summed to obtain word embeddings for downstream tasks. Since these approaches deterministically reduce the sequence length in the downsampling operation, they may not capture the morphological information, potentially struggling to learn representations on noisy text.

## 3 Method

We propose a learnable tokenizer that is trained to convert sequence of characters into meaningful subword-level tokens. Consider the multilingual alphabet $\Psi$ (i.e., a closed set of letters) and the character sequence $c = [c_1, \ldots, c_n]$ that represents a word of length $n$ and $c_i \in \Psi$. We aim at learning the corresponding IOB[1] sequence of tags $t = [t_1, \ldots, t_n]$ that groups characters into the desired tokenization:

$$p_\theta(t \mid c, \ell) = f_\theta(c, \ell) \qquad (1)$$

Here $\ell$ denotes the language of the word. The model $f_\theta$ can be any neural architecture that allows a one-to-one mapping from the input to the output.[2] We condition the model on $\ell$ in the multilingual setting, while the monolingual variant does not require it.

A trained neural tokenizer, $f_\theta$, is capable of providing tokenization as a stand-alone tool, which can be compared directly to the standard subword tokenizers (e.g., controlling by the task-specific model in a downstream setting). Additionally, a trained neural tokenizer can expose the internal representations of a segmentation so that it enables end-to-end task learning by optimizing the tokenization towards the task particularities. We describe both scenarios in more detail in the following subsections.

### 3.1 Pre-training

We rely on the assumption that statistical subword tokenizers learn reasonable tokenization until they start over-segmenting the text due to the target vocabulary size and the infrequent subword occurrences. To stick to a data-driven approach (hence, avoiding language specific heuristics), we choose a subword tokenizer, i.e. Unigram (Kudo, 2018), to generate our ground-truth segmentation while also discarding over fragmented sequences. For example, if the subword tokenizer segments **tricycles** as **tri/cycle/s**, then the ground-truth label is `BIIBIIIB`. We train our neural tokenizer using the negative log-likelihood objective over the subword tokenizer segments:

$$\mathcal{L} = -\sum_i t_i \log p_\theta(t_i \mid c, \ell) \qquad (2)$$

---
[1] The beginning (B), inside (I), and outside (O) tagging schema denoting the word boundaries at the character level.
[2] We stick to the LSTM architecture for all our experiments since this simplifies iterations over pre-training and fine-tuning.

Our neural tokenizer not only mimics the more prominent (and insightful) patterns from the subword tokenizer, but it also generalizes such behaviors to unseen words.

**Pre-training Dataset:** We generate a pre-training corpus by curating space-separated tokens from the Wikipedia articles (e.g., removing hyperlinks, HTML tags, and tokens whose lengths are beyond 30 characters). Additionally, we use two heuristics to improve the ground-truth label from the subword tokenizer. First, if the input sequence is less than four, we do not segment into subwords. Second, if the subword tokenizer creates more than 50% subwords with a single character, we discard the ground-truth label and do not tokenize. These heuristics discard the junk tokenization of the subword tokenizer, especially when the input comes from low-resource languages and noisy text.

### 3.2 End-to-End Task Learning

While the pre-training provides a stand-alone neural tokenizer tool, we can also leverage the model's hidden representations for end-to-end task learning. Recall that the neural tokenizer provides a tagging sequence for the segmented tokens based on its internal character-level vectors. Such tags can be used to group and reduce the dimensionality of the internal representations (e.g., via max-pooling). Our neural tokenizer is based on the LSTM architecture, so we use the LSTM output vectors and max-pool them according to the tokenization tags (although this approach is invariant to LSTM).

$$[h_1, \ldots, h_n] = \text{LSTM}([c_1, \ldots, c_n]) \qquad (3)$$
$$r_i = \text{maxpool}([h_i, \ldots, h_j])$$

where the interval $[i, j]$ denotes the characters of a single subword (i.e., an IOB segment), $h_i \in \mathbb{R}^{1 \times d}$ and $r_i \in \mathbb{R}^{1 \times d}$ are vectors of dimensionality $d$. We use the resulting vectors $r$ as the subword representations, which we can feed to any task-specific model on a downstream scenario. Note that we effectively bypass the need of a vocabulary, while also enabling the task-specific model to adjust the pre-trained tokenization parameters towards the task domain in an end-to-end manner.

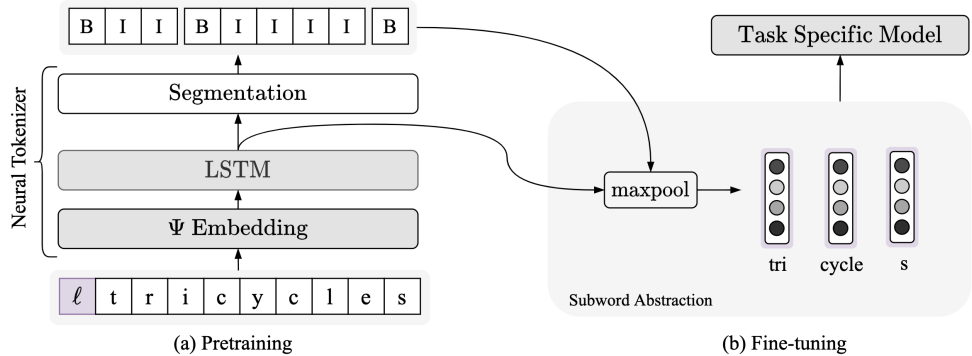

Figure 1: The neural tokenizer architecture and its two settings: (a) Pre-training and (b) Fine-tuning. (a) In the pre-training setting, the model is trained to segment the sequence of characters by outputting the correct IOB tags according to the statistical subword tokenizer. (b) In the fine-tuning setting, the model uses the trained segmentation layer to predict the tags and max-pool the corresponding vectors (e.g., tri/cycle/s). These vectors are passed directly to the task-specific model, bypassing the need for vocabulary and embedding layers. In the backpropagation step of the fine-tuning setting, all the parameters in the shadow boxes are updated (i.e., the alphabet embedding, LSTM, and the task-specific parameters).

### 3.3 Neural Tokenizer Variants

The neural tokenizer model can be used to segment the input for a task model in a general-purpose setting, such as a task with monolingual input (i.e., $\ell$ is constant). However, we need to slightly change the neural tokenizer model for multilingual and mixed-lingual (code-switched) settings to improve the tokenization and internal representations. We describe two variants of our neural tokenizer: *multilingual* and *mixed-lingual* neural tokenizers.

**Multilingual Neural Tokenizer:** Multilingual subword tokenizers are designed to segment the text with an fixed multilingual vocabulary, irrespective of the input language. While this may be practical, it has severe effects on the tokenization behavior, disregarding dissimilar linguistic properties across languages (e.g., morphology). Thus, if a language identifier $\ell$ is available with the input, the neural tokenizer can condition the tokenization on $\ell$. We achieve such behavior by simply including the identifier $\ell$ at the beginning of the sequence, which extends the alphabet $\Psi$ with the same number of languages $\ell$ we are including in the pre-training data.

Additionally, since multilingual subword tokenizers cannot tokenize low-resource languages appropriately due to the dominance of the high-resource languages in their vocabulary, we use monolingual subword tokenizers to generate the ground-truth segmentation labels for pre-training. Using monolingual subword tokenizers helps our neural tokenizer avoid bias towards any languages, especially the high-resource languages. Thus, we distill the tokenization knowledge from the multiple monolingual subword tokenizers into our neural tokenizer.

**Mixed-Lingual Neural Tokenizer:** In mixed-lingual settings, such as in code-switching, we may not have access to the language identifiers $\ell$ of the input words or sentences. Thus, we need to train a neural tokenizer to segment text with mixed languages without relying on language identifiers of the input tokens. To do so, we change the pre-training dataset to train the neural tokenizer with and without language tags, hence forcing our model to generalize when the language tags are provided as well as when they are missing. We replicate the dataset for training the model with and without language tags.

## 4 Experimental Setup

### 4.1 Neural Tokenizer Model

We design the neural tokenizer character encoder with a character embedding layer of 64 dimensions followed by a two-layer bidirectional LSTM (Bi-LSTM) that generates 128-dimensional vectors. We use a fully connected layer of shape $128 \times 2$ followed by a softmax operation to predict the character-level segmentation label. The predicted labels represent whether a character is the beginning or part of a subword.[3]

### 4.2 Pre-training Neural Tokenizer

In the pre-training phase of the monolingual neural tokenizer, we have developed a monolingual Unigram subword tokenizer with a vocabulary size of 30, 000 to generate the ground-truth segmentation labels. To train multilingual and mixed-lingual (code-switched) neural tokenizers, we have developed monolingual Unigram tokenizers with a vocabulary of 30, 000 for each language. We fixed the vocabulary size by following monolingual vocabulary size of BERT (Devlin et al., 2019).

---
[3] Note that the O tag of the IOB schema is not used here.

We have utilized Adam optimizer with weight decay regularization and cosine annealing warm restarts with an initial learning rate set to $3e^{-4}$ to train the neural tokenizer. In the cosine annealing warm restarts learning scheduler, we set the cycle length ($T_0$) and cycle multiplier ($T_{mult}$) to 3 and 2, respectively. We have trained the models for 6 epochs and selected the best model based on the minimum validation loss.

### 4.3 Baseline Tokenizers

We have developed subword tokenizers, such as BPE, Unigram, and Word-Piece, for the experimental evaluations. We developed two versions of these subword tokenizers: monolingual and multilingual. Following state-of-the-art model with subword tokenizer (Devlin et al., 2019), we have fixed the vocabulary size of monolingual and multilingual tokenizers to 30000 and 120000, respectively. We have used the Wikipedia dataset to develop the vocabulary of these tokenizers.

### 4.4 Downstream Task Model

We have evaluated the impact of our neural and heuristic-based subword tokenizers on the two downstream tasks: natural language inference (NLI) in monolingual and multilingual settings and sentiment analysis with code-switched languages. For the baselines models with subword tokenization, the segmented subwords are projected to create feature embeddings of size 256. For the model with our neural tokenizer, we max-pool the character embeddings to create the subword-level representations of size 128. We project these pooled representations to the embeddings of size 256 to match the subword representations' dimension of the baseline tokenizers. We have used a two-layers Bidirectional LSTM with the hidden feature embeddings of size 256 for extracting the task representations. In the experimental evaluations, we have used the same task model architecture with the subword tokenizer and our neural tokenizer. All the models are trained from scratch for fair experimental evaluations.

## 5 Experimental Results and Discussion

We have evaluated the impact of our neural and subword tokenizers on multilingual and monolingual natural language inference (NLI) tasks and on a sentiment analysis task with code-switched language. We also evaluated the impact of tokenizers in the presence of noisy data (typos and misspelling).

### 5.1 Evaluations on Multilingual NLI Tasks

We have conducted the experimental analysis to evaluate the impact of neural and baseline tokenizers on multilingual NLI tasks with five languages: Arabic (ar), English (en), Russian (ru), Swahili (sw), and Thai (th). We have used XNLI dataset (Conneau et al., 2018) for this experimentation. We have developed three multilingual tokenizers (BPE, Unigram, and Word-piece) with a vocabulary size of 120, 000. Moreover, we have developed another baseline, called Character-based Model, which segments input based on space without using any vocabulary and pools character embedding to create word-level representations. These representations are used for downstream task. Finally, we have used the same downstream learning architecture (Described in Section 4.4) with all the above-mentioned tokenizers and multilingual neural tokenizers.

**Results and Discussion:** The experimental results in Table 2 suggest that the neural tokenizer outperforms the evaluated baseline tokenizers across all languages for the NLI task. Especially, neural tokenizer achieves substantially larger gains for the low-resource languages over the baseline tokenizers, such as +11 absolute points of accuracy for Thai (th), +8 for Arabic (ar), and +4 for Swahili (sw). For the English, neural tokenizer slightly improves the performance compared to the baseline tokenizers.

The reasoning behind the performance improvement of neural tokenizer is that it segments the input based on lexical similarity and thus create better segmentations, especially for the low-resource languages. As the subword tokenizers use a vocabulary with the most frequent subwords in a corpus, these tokenizers over-segment the input of low-resource languages and create junk tokens, which lead to the performance degradation.

We have also noticed similar phenomena in our qualitative analysis, presented in Figure 2 and 3. Subword tokenizers over-segment the words from the low-resources languages compared to the high-resources languages (Figure 2). For example, the subword tokenizers create at least 10 subwords for more than 20% words in the multilingual NLI corpus. On the other hand, the neural tokenizer creates fewer subwords than the baseline tokenizers, including the Unigram, which is used to pre-train our neural tokenizer. Specifically, the neural tokenizer reduces the number of subwords for the low-resource languages, such as Thai (th) and Swahili (sw). As the neural tokenizer distills the segmentations knowledge from the language-specific tokenizer, it does not bias towards the high-resource languages. Additionally, we have observed that subword tokenizer over-segment the hypothesis and premise from low-resource languages, such as Arabic (ar), Swahili (sw), and Thai (th), compared to the neural tokenizer (Figure 3). This over-segmentation leads to performance degradation for the NLI task with low-resource languages.

Table 2: Multilingual NLI task performance comparison of various tokenization approaches.

| Tokenizers | Vocab Size | Model Params (Millions) | Languages (Accuracy %) | | | | |
|---|---|---|---|---|---|---|---|
| | | | ar | sw | th | ru | en |
| BPE | 120,000 | 67.8 M | 51.81 | 50.66 | 51.32 | 54.77 | 57.57 |
| Unigram | 120,000 | 67.8 M | 53.78 | 51.32 | 56.09 | 53.13 | 57.24 |
| Word-Piece | 120,000 | 67.8 M | 50.66 | 50.00 | 43.26 | 54.61 | 57.57 |
| Character-based Model | - | 33.4 M | 53.29 | 46.88 | 44.41 | 52.80 | 50.99 |
| **Neural** | - | **33.4 M** | **61.51** | **53.95** | **68.42** | **60.69** | **58.22** |

Additionally, one can argue that instead of using the neural tokenizer, we can use a Character-based Model to extract characters embedding for downstream task learning. To validate this argument, we have developed a baseline, called Character-based Model, which segments input based on space without using any vocabulary and pools character embedding to create word-level representations for downstream task learning. This Character-based Model is trained end-to-end to learn characters embedding from input character sequence and generate task representation to produce task output. The results in Table 2 suggest that although it achieves comparable performance to the baseline subword tokenizers, there is a considerable performance gap between the Character-based Model and the neural tokenizer across all the languages.

Moreover, our neural tokenizer achieved these performance improvements with half the model size compared to the model with baseline tokenizers. Because the model with baseline subword tokenizer has to allocate most of the model parameters to learn the subword embeddings. On the other hand, neural tokenizer creates the subword embeddings by pooling the character-level representations, which reduces the model size.

Table 3: Monolingual (English) NLI task performance comparison with various tokenization approaches

| Tokenizer | Vocab Size | Model Params (Millions) | Accuracy (%) |
|---|---|---|---|
| BPE | 30,000 | 44.8 M | 57.85 |
| BPE | 70,000 | 65.3 M | 59.94 |
| Unigram | 30,000 | 44.8 M | 58.65 |
| Unigram | 70,000 | 65.3 M | 58.01 |
| Word-Piece | 30,000 | 44.8 M | 58.65 |
| Word-Piece | 70,000 | 65.3 M | 58.33 |
| **Neural** | 69,480 | 65.0 M | 59.94 |
| **Neural** | - | **33.3 M** | **60.58** |

## 5.2 Experimental Evaluations on Monolingual NLI Tasks

We have investigated whether the neural tokenizer can outperform the baseline subword tokenizers on monolingual NLI tasks. We have developed three baseline subword tokenizers (BPE, Unigram, and Word-Piece) with the vocabulary of sizes 30,000 and 70,000. To ensure a fair comparison, we have also trained our neural tokenizer in the monolingual setting. Moreover, we have applied our neural tokenizer on the same corpus as the baseline tokenizers and produced a vocabulary for the neural tokenizer. In this vocabulary-based setting, we tokenize the input based on the fixed vocabulary, similar to baseline subword tokenizers. In this experimental evaluation, we have selected the English language.

**Results and Discussion:** The experimental results in Table 3 suggest that our neural tokenizer shows comparable performance to the subword tokenizers on the monolingual NLI task. Moreover, the model with our neural tokenizer achieves a similar performance to the model with the subword tokenizers. However, the neural tokenizer helps to achieve similar performance with a smaller model. This performance improvement of neural tokenizers with reduced model size attributes that we can utilize our neural tokenizer to extract representations for downstream tasks instead of employing a resource-intensive model with subword tokenizers.

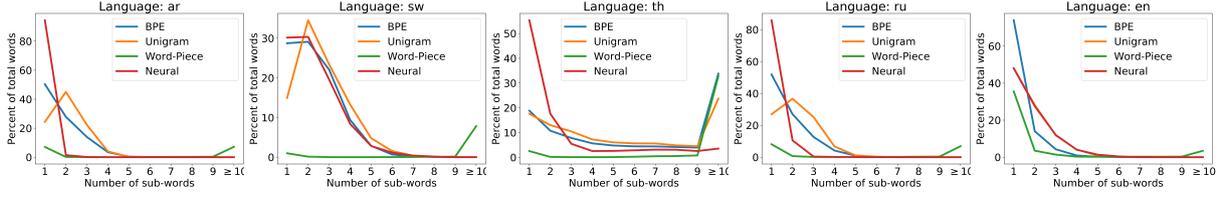

Figure 2: Impact of tokenizer to segment words into different number of subwords in low and high resource languages.

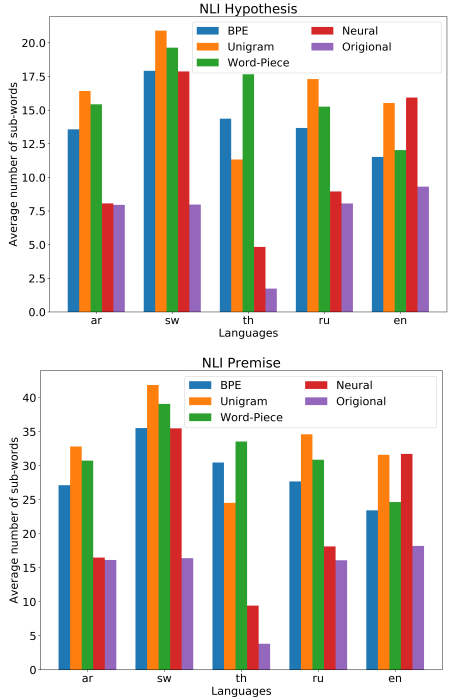

Figure 3: Average number of subwords of hypothesis and premise from low and high resource languages, which are tokenized by different tokenizers.

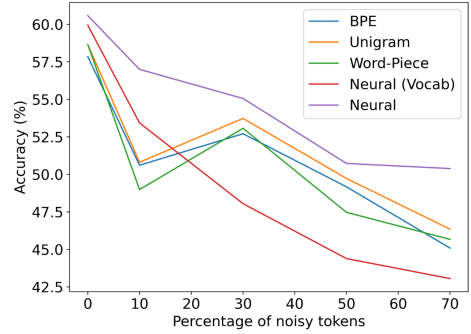

(a) Accuracy of NLI (English) Task

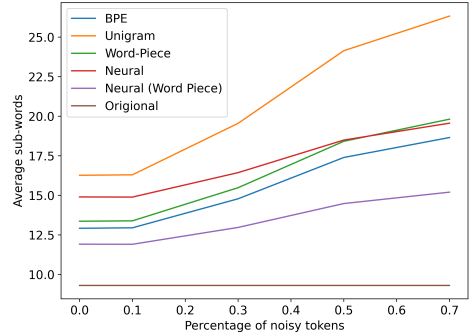

(b) Average number of segmented subwords

Figure 4: Performance comparison of tokenizers on monolingual (English) NLI task with noisy text (typos and misspelling).

### 5.3 Experimental Evaluations on Noisy Text

We have evaluated the impact of tokenizer on monolingual (English) NLI task in the presence of noisy text (typos and misspelling). For this experimental evaluation, we have developed baseline monolingual (English) tokenizers (BPE, Unigram, and Word-Piece) with a vocabulary size of 30,000. We have developed a monolingual neural tokenizer trained, which is trained using a Unigram subword tokenizer with a vocabulary size of $30,000$. Moreover, we have developed a vocabulary-based neural tokenizer, where we used our neural tokenizer to segment the Wikipedia corpus with the English language and create a vocabulary with the most frequent subwords. We have used this vocabulary to tokenize the hypothesis and premise of the NLI task. We adversarially introduce noise to the $0\% - 70\%$ input words, such as typos and misspelling, using TextAttack Library (Morris et al., 2020).

**Results and Discussion:** The experimental results in Fig 4 suggest that the performance of the models with vocabulary-based subword tokenizers degrade with the increased amount of noisy words in the input. Although the performance of the model with our vocabulary-free neural tokenizer degrades with the increased amount of noisy words, it outperforms all the evaluated tokenization approaches. Especially, our neural tokenizer outperforms the Unigram subword tokenizer, which is used to train our neural tokenizer.

As the subword tokenizers use a fixed vocabulary, they can not appropriately segment the text from the out-of-distribution and introduce junk tokens. These junk tokens lead the model to create suboptimal representations and thus degrade the downstream task's performance. On the other hand, neural tokenizer segments the input based on lexical similarity, and thus it creates better segmentation in the presence of noise, such as typos and misspelling. However, the vocabulary-based neural tokenizer's performance degrades with the increased percentage of noisy words. Because, in the vocabulary-based neural tokenization, if a subword does not present in the vocabulary, then we replace that subword with an *<UNK>* (unknown) token. As a result, vocabulary-based neural tokenizers create many *<UNK>* junk subwords in the presence of noise and thus hurting the task performance.

Table 4: Segmentation of words using Unigram and Neural Tokenizer, which is trained using Unigram subword tokenizer. Red colored words are with noise (typos and misspelling).

| Input | Unigram | Neural |
|---|---|---|
| tricycles | t/ r/ i/ cycle/ s | tricycle/ s |
| trycycles | t/ r/ y/ cycle/ s | trycycle/ s |
| improving | improv/ ing | improv/ ing |
| improbing | imp/ robin/ g | improbing |
| timeline | timeline | time/ line |
| timlline | t/ i/ m/ l/ line | timlline |
| swimming | s/ w/ imming | s/ w/ imming |
| swiming | swim/ ing | swiming |
| workshop | workshop | workshop |
| worksops | works/ o/ p/ s | worksops |
| biotechnology | biotechnolog/ y | biotechnolog/ y |
| bitechnology | b/ i/ t/ echnology | bitechnolog/ y |

Additionally, the tokenizations present in Table. 4 suggest that our neural tokenizer helps to improve the segmentations quality of the Unigram subword tokenizer in the presence of noise (e.g., typos and spelling variations). For example, Unigram creates junk tokens in segmenting *tricycles*. Our neural tokenizer, trained using Unigram, reduces the junk tokens and creates morphologically aligned segmentation. However, in some cases, such as segmenting swiming, the Unigram tokenizer creates better segmentation than our neural tokenizer.

### 5.4 Experimental Evaluations on Code-Switched Language

We have evaluated the impact of the tokenizers on the sentiment analysis task with the Spanish-English code-switched languages. We have used the Lince dataset and the evaluation benchmark (Aguilar et al., 2020a). We have developed three baseline tokenizers (BPE, Unigram, Word-Piece) with a vocabulary size of 60,000. We have also trained a neural tokenizer in the mixed-lingual settings (Section 3.3), where the training dataset is developed from the Spanish and English Wikipedia articles.

**Results and Discussion:** The experimental results in Table 5 suggest that our neural tokenizer outperforms the subword tokenizers, including the Unigram subword tokenizer, on the sentiment anal-ysis task with code-switched languages. Unlike the heuristic-based subword tokenization, neural tokenizer allows end-to-end task learning, which helps to improve the task's performance.

Table 5: Performance comparison of tokenizers on sentiment analysis task with code-switched languages (Spanish-English).

| Tokenizer | Vocab Size | Accuracy (%) |
|---|---|---|
| BPE | 60,000 | 49.39 |
| Unigram | 60,000 | 49.18 |
| Word-Piece | 60,000 | 48.43 |
| Character-based Model | - | 45.63 |
| **Neural** | - | **51.41** |

Our neural tokenizer and the heuristic-based tokenizers segment the input into subwords, and the task models use the subword embeddings. These models, which use the subword embeddings, outperform the Character-based Model, where character representations are used for downstream task learning. Because in the code-switched language settings, extracting subword embeddings can be beneficial to create aligned multilingual representations, which help to improve the sentiment analysis task performance. Thus, appropriately segmenting input with code-switched languages is crucial to improve performance in the code-switched language settings.

## 6 Conclusion

We propose a neural tokenizer to segment text without a vocabulary, which allows end-to-end task learning. The experimental evaluations on multilingual NLI task suggest that our neural tokenizer reduces the model size and improves the task's performance for low-resources languages, such as Arabic, Swahili, and Thai. Moreover, the neural tokenizer outperforms subword tokenizers on the NLI task with noisy text (typos and misspelling). The qualitative analysis also suggests that our neural tokenizer improves the tokenizations of the subword tokenizers, which is used to train our neural tokenizer. Additionally, the neural tokenizer shows comparable performance on sentiment analysis task with code-switched languages. The experimental results suggest that our neural tokenizer can distill the segmentations knowledge from multiple subword tokenizers to improve the tokenization. This finding opens future research avenues to design a learnable tokenizer for improving the state-of-the-art subword tokenization and the downstream task's performance.